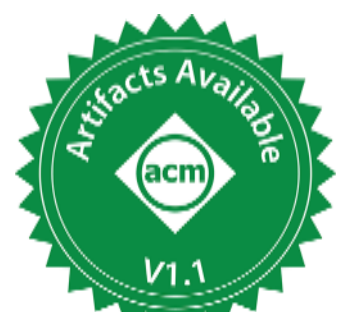

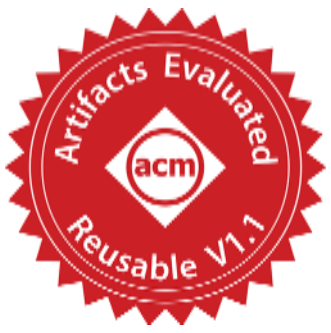

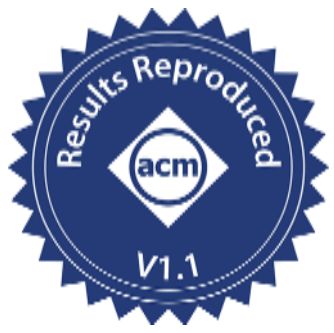

# Using (Not-so) Large Language Models to Generate Simulation Models in a Formal DSL: A Study on Reaction Networks

JUSTIN NOAH KREIKEMEYER, Faculty of Computer Science and Electrical Engineering, Institute for Visual and Analytic Computing, University of Rostock, Rostock, Germany
MIŁOSZ JANKOWSKI, Faculty of Computer Science and Electrical Engineering, Institute for Visual and Analytic Computing, University of Rostock, Rostock, Germany
PIA WILSDORF, Faculty of Computer Science and Electrical Engineering, Institute for Visual and Analytic Computing, University of Rostock, Rostock, Germany
ADELINDE M. UHRMACHER, Faculty of Computer Science and Electrical Engineering, Institute for Visual and Analytic Computing, University of Rostock, Rostock, Germany

Formal languages are an integral part of modeling and simulation. They allow the distillation of knowledge into concise simulation models amenable to automatic execution, interpretation, and analysis. However, the arguably most humanly accessible means of expressing models is through natural language, which is not easily interpretable by computers. Here, we evaluate how a Large Language Model (LLM) might be used for formalizing natural language into simulation models. Existing studies only explored using very large LLMs, like the commercial GPT models, without fine-tuning model weights. To close this gap, we show how an open-weights, 7B-parameter Mistral model can be fine-tuned to translate natural language descriptions to reaction network models in a domain-specific language, offering a self-hostable, compute-efficient, and memory efficient alternative. To this end, we develop a synthetic data generator to serve as the basis for fine-tuning and evaluation. Our quantitative evaluation shows that our fine-tuned Mistral model can recover the ground truth simulation model in up to 84.5% of cases. In addition, our small-scale user study demonstrates the model's practical potential for one-time generation as well as interactive modeling in various domains. While promising, in its current form, the fine-tuned small LLM cannot catch up with large LLMs. We conclude that higher-quality training data are required, and expect future small and open-source LLMs to offer new opportunities.

The authors acknowledge the funding of the Deutsche Forschungsgemeinschaft under Grant No.: 320435134 (https://gepris.dfg.de/gepris/projekt/320435134).
Authors' Contact Information: Justin Noah Kreikemeyer, Faculty of Computer Science and Electrical Engineering, Institute for Visual and Analytic Computing, University of Rostock, Rostock, MV, Germany; e-mail: justin.kreikemeyer@uni-rostock.de; Miłosz Jankowski, Faculty of Computer Science and Electrical Engineering, Institute for Visual and Analytic Computing, University of Rostock, Rostock, MV, Germany; e-mail: milosz.jankowski@uni-rostock.de; Pia Wilsdorf, Faculty of Computer Science and Electrical Engineering, Institute for Visual and Analytic Computing, University of Rostock, Rostock, MV, Germany; e-mail: pia.wilsdorf@uni-rostock.de; Adelinde M. Uhrmacher, Faculty of Computer Science and Electrical Engineering, Institute for Visual and Analytic Computing, University of Rostock, Rostock, MV, Germany; e-mail: adelinde.uhrmacher@uni-rostock.de.

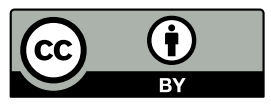

## 1 Introduction

Creating simulation models is a knowledge-intensive process, heavily relying on the expertise of domain experts [53]. Depending on the domain (systems biology, ecology, sociology, manufacturing, etc.), experts may draw on **domain-specific languages** (**DSLs**) that make formal modeling more accessible to them [55]. Still, to effectively use a (new) language or formalism, the modeler has to become acquainted with its syntax and semantics. Formulating the system's dynamics using natural language would make it possible to skip or facilitate this process. Also, in the case of automatically generating simulation models [38, 55], it would open up an additional source of information.

Natural language is arguably the most accessible and intuitive means for humans to express knowledge about a system to be modeled, as evidenced by decades of research in **natural language processing** (**NLP**) [10]. Of course, the translation from a natural language to a formal one is prone to be incomplete and ambiguous: a natural language, as opposed to a formal language, allows for missing information, varying interpretations, or contradictions [16]. Still, the approximate solution to this problem is highly relevant, and the present paper is hardly the first to tackle the question: To what extent can modeling with DSLs be supported by automatically translating from natural language to a formal model description? [2, 9, 13] (cf. also Section 3).

Considering the recent advances in NLP with **large language models** (**LLMs**), it is of great interest to evaluate their potential role in solving the formalization task. Current LLMs are neural networks with around $10^9$ to $10^{12}$ parameters, typically following a transformer architecture [56] (cf. Section 2.2). By training them on massive amounts of (curated) text data, they become accurate generative models of natural language. Further, they can be fine-tuned on conversational-style or instruction-style data, so they model conversations with humans and can accept their commands.

Several papers already discuss using LLMs for the model translation task, aiming at specific use-cases like logistics [31] or biochemistry [44]. However, they rely solely on the generalization capabilities of very large models. This is done by engineering the right "prompt" and examples for an LLM to prepare it for the specific task at hand, commonly referred to as zero-shot or few-shot learning. Despite already achieving passable results, they leave the use of more advanced training, such as parameter-efficient fine-tuning and other techniques (cf. Section 4) unexplored.

*We argue that a combination of fine-tuning, few-shot prompting, and other techniques allow the use of much smaller, open-weight LLMs for the translation task, requiring only a fraction of the resources and providing full control over reproducibility.*

As a case study, we investigate the formalism of **Chemical (kinetic) Reaction Networks** (**CRNs**), which is used to describe interactions between entities in a system and the resulting changes in their populations over time [25, 28]. Despite their name, they are used throughout various domains and systems where the dynamics can be adequately expressed through changes in populations, e.g., systems biology, ecology, and epidemic modeling. In this article, we fine-tune and evaluate an LLM to generate CRNs from natural language descriptions of the involved reactions. The goal is not to train the LLM on expert knowledge from specific domains, but rather to enable it to reliably translate the jargon of different domains into a more formal, machine-interpretable description. In other words, understanding how experts talk will ultimately enable LLMs to translate the domain knowledge as formulated by the expert into a simulation model.

To this end, we explore the generation of synthetic training data (Section 4.2), the use of few-shot learning (Section 4.3), **parameter-efficient fine-tuning** (**PEFT**) [62] with **low-rank adaptation** (**LoRA**) [30] (Section 4.4), and constrained decoding to enforce adherence to a formal language's syntax (Section 4.5). We start by introducing the concepts of population modeling with CRNs and NLP with LLMs in Section 2 and reviewing the related work in Section 3. Section 4 introduces the methods used in this article, as outlined above, in more detail. In Section 5, we prepare an open-weights, seven-billion-parameter LLM by Mistral for translating natural language to CRNs and extensively evaluate its performance when using the different methods. We also compare it to the commercial GPT-4o model [15], which has orders of magnitude more parameters [39], the existing KinModGPT approach, and the INDRA tool, which uses traditional reading engines. The evaluation is concluded with a small-scale user study, which provides insights into current limitations and reveals important avenues for improvements.

## 2 Background

This article bridges two distinct fields, bringing together dynamic population models from simulation modeling and LLMs from NLP. In the following, we will summarize and connect both topics.

### 2.1 Population Modeling with Reaction Networks

Modeling is the process of finding a useful abstraction to describe an observed phenomenon or system. In many cases, the system of interest involves individual entities (agents) and their interactions, leading to agent-based models. The *population-based* paradigm further abstracts from this setting by only considering the total population counts of the involved agent types (species). This approach is particularly suitable if the individual entities can easily be grouped by their properties and relationships, as is often the case in, e.g., epidemiology [7]. One well-known formalism used to express (and interpret) population models is as a set of **ordinary differential equations** (**ODEs**). In many cases, population models are described as a set of *coupled* ODEs, where terms are shared between the derivative of each species (e.g., an increase in the population or concentration of infected persons $dI/dt = \alpha SI - \beta I$ leads to an equivalent decrease of the population of susceptible persons $dS/dt = -\alpha SI$).

The formalism of CRNs acknowledges this fact by making interactions and their results explicit in the form of reactions, e.g., $S+I \xrightarrow{\alpha \cdot \#S \cdot \#I} 2I$ with $\#X$ being the size of the population of the species $X$. CRNs can be easily mapped (on demand) to different semantics for execution [6], e.g., being interpreted deterministically as a set of ODEs, stochastically as a **Continuous-Time Markov Chain** (**CTMC**) or a hybrid of both.

A CRN with $n$ reactions over $m$ species $S_j, j = 1, \ldots, m$ is a set

$$\left\{ R_i : \sum_{j=1}^{m} a_{ij} S_j \xrightarrow{r_i} \sum_{j=1}^{m} b_{ij} S_j \;\middle|\; i = 1, \ldots, n \right\}, \tag{1}$$

where the coefficients $a_i$ define the reactants (left side) and the coefficients $b_i$ define the products (right side) of the reaction. Commonly, reactants/products with coefficient 0 and coefficients of value 1 are omitted in a model description. The speed at which the reaction is executed is governed by a rate constant $r_i$. When a reaction executes (fires), the reactants are subtracted from, and the products are added to the populations. For example, the reaction $S + I \xrightarrow{0.1} 2I$ would decrease the population of $S$ by one and increase the population of $I$ by one at a rate (assuming mass action kinetics [17]) of $0.1 \cdot \#S \cdot \#I$.

Based on the CRNs, various *rule-based* modeling approaches have been introduced [11]. These allow the definition of multiple reactions with a single rule by introducing placeholders, e.g., $A(free, *) + B(free) \longrightarrow A(bound, *) + B(bound)$; this rule is applied to all $A$s independently of their other attributes.

Although DSLs cater to the need of the modeler for succinct modeling and try to keep the entrance barrier for modelers low by, e.g., building on a familiar modeling metaphor, such as chemical reactions, each new DSL still needs to be learned. The modeler has to acquire its specific syntax and familiarize with its ideas (and semantics). This might be mitigated by using NLP to translate a natural language formulation to different DSLs, thus supporting the modeler in learning a new language, applying it, or even partly replacing the efforts required by the modeler.

### 2.2 Natural Language Processing with Large Language Models

In the past 7 years, major breakthroughs have been made in NLP [14, 20, 50, 56]. Probably the most important was the inception of the transformer architecture, which forms the basis for most of today's LLMs [20]. It combines an encoder-decoder neural network architecture (today, often only the decoder is used) for sequence processing and a mechanism for weighing atomic parts of the input sequence, called *attention* [56]. Intuitively speaking, the latter enables the model to determine the context of (even distant) words in a text. Trained on vast amounts of (text) data, LLMs achieve unprecedented performance on generating natural-sounding texts and generalization to many other NLP tasks that previously required separate approaches [14, 20]. By fine-tuning such a foundational generative model on conversation-style data (e.g., conversations between an "assistant" and a "user"), they have even been extended to act as an assistant to a human end-user [50].

For detailed information on the architecture of LLMs, we refer the reader to, e.g., [14, 56]. Here, we will give a rough overview of the most important concepts when working with existing pre-trained models like ChatGPT [15] or Mistral [33]. The first step in the language generation process with a transformer model is transforming the input letter sequence, called the *prompt*, into a sequence of pre-defined substrings, called *tokens*. Special embedding models [48] are then used to assign a corresponding vector to each token. Transformer models, based on their architecture, can often only accept and attend to a maximum number of input vectors, which is called their *context length*. These vectors are then provided as input to the (trained) transformer, which predicts a probability distribution over the next token to follow. A softmax function, scaled by the *temperature* parameter $t$, is used to map the output logits from the neural network to a probability distribution. The distribution over all output sequences would have to be sampled to get the most likely completion. As the number of possible tokens to choose from grows exponentially with the output sequence's length, different strategies need to be used to sample from this distribution, e.g., by only considering the current token distribution. A popular one is to use *beam-search*, which retains a set of $k$ greedily selected completions. Instead of selecting the token(s) with the highest probability, the decoding is typically randomized based on the probabilities. This process can be steered by the softmax function's temperature parameter, which scales the selection from almost deterministic ($t \rightarrow 0$) to uniformly random ($t \rightarrow \infty$) by shaping the distribution built from the logits accordingly. The last step is to map the decoded token sequence back to text.

A trained transformer is fully specified by its neural network weights and biases, often just called its weights or parameters. Highly parametrized versions of transformers, like GPT [50], Llama [41] or Bert [14], referred to as LLMs, generalize to perform many NLP tasks previously solved by separate approaches [20]. Besides high-quality training data, the number of parameters seems to be a key ingredient in this success. However, increasingly smaller models are shown to achieve performance close to their larger siblings with much less computational and memory resources [33]. These not-so-LLMs achieve this by curating even better datasets, distilling the knowledge of larger models, or using new developments in model architecture, like sliding window attention [33]. With these developments also come efforts to develop models with openly licensed weights (open weights) or training procedure and data (open source) that allow for broad application and reproducible research, which highly parametrized commercial LLMs often do not.

## 3 Related Work

Especially since their popularization by ChatGPT [50], there has been a massive surge of interest in research on solving NLP tasks with LLMs [20]. A focus of many efforts lies on achieving a natural interaction between humans and computer programs, e.g., searching document databases with natural language search queries and synthesizing answers [40]. Consequently, translating from natural language requests or descriptions to a formal output, e.g., program code in a general-purpose programming language, also received a lot of attention [4].

One line of research focuses on supporting this process by constraining the output of LLMs to a formal specification [5, 23, 58], e.g., an **Extended Backus-Naur Form** (**EBNF**) or other formal grammar specification. To this end, **grammar-constrained decoding (GCD)** can be used to prune the probability distribution over tokens proposed by the LLM to only those tokens that are valid continuations of the existing string according to a context-free grammar [23]. In [58] "grammar prompting" is proposed to augment each query with a specialized subset of the full BNF specification of a complex DSL. The authors of [5] test a combination of **retrieval-augmented generation (RAG)** to select fitting few-shot examples for a given query. As in [58], this provides the model with specific and relevant examples to answer the current user instruction.

Further, several works propose to generate simulation models based on natural language descriptions using LLMs, for example, for biochemical reaction models [44], agent-based models [9], and logistics models [31]. In [9], an assistant is designed as a learning environment for the NetLogo agent-based modeling language [59] using GPT-3.5 Turbo as a chat backend and prompt engineering. In [46], NetLogo models are generated using GPT-3.5 Turbo through two other strategies: few-shot prompting and RAG to incorporate additional examples from an external database. Others do not generate code but integrate the LLM into the simulation model, e.g., to mimic human-like reasoning and decision-making of agents [22]. The work presented in [44] is probably the closest to our paper, aiming to determine a CRN in the SBML [36] format from its natural language description. However, all the above approaches only consider using very large commercial models, like ChatGPT [50], without tuning their weights. The evaluation is often based on only a couple of larger, hand-picked examples. A larger-scale evaluation is thus an interesting research direction to pursue. In addition, we here investigate the use of methods to prepare considerably smaller, open-weights models for the translation to a DSL and extensively evaluate the capabilities of the resulting models. Similar challenges are faced when generating code for the class of low-resource (lacking large amounts of data), domain-specific programming languages—which, of course, includes modeling languages. [34] survey methodologies for enhancing LLM performance (incl. fine-tuning, prompting, decoding, and early stopping), as well as methods for the curation of data sets and the use of pre-trained models of different sizes across a variety of languages. Overall, [21] postulate that "The era of ModelGPT or SimulationGPT is coming", and discuss the potentials of a "modeler in a box" assistant for agent-based modeling and modeling in systems dynamics. One particular benefit of LLM-based approaches is that they can also broaden access to simulation end-users by helping select and generate relevant models for the research question without profound modeling experience [24].

Extracting formal models from natural language descriptions and literature has been a research topic long before the advent of LLMs [13, 54]. This line of research also continues to this day. Systems biology is an example of an application domain of modeling and simulation where automation plays an increasing role and is actively researched [8, 26]. For example, the ACCORDION tool [2] can extract biological networks from literature databases and texts. This and similar tools use so-called reading engines (e.g., TRIPS [3]) to extract information about the nodes and edges of a reaction network graph. Similarly, the INDRA tool [27] was designed to extract formal models from biological word models. In [26], a classification of approaches according to their degree of automation is given; our work fits into level two, where the modeling is fully automated, but the modeler still provides natural language expert input.

## 4 Methods

In this section, we describe the existing methods we combine to adapt a pre-trained LLM to the task at hand: Translating from a natural language description to a formal language model. This involves (1) obtaining training data with examples for correct translations, (2) providing the model with those examples, either by few-shot prompting or adjusting the model weights, and (3) optionally constraining the generation to produce only valid continuations according to a formal grammar.

### 4.1 Description of Task

We begin with a detailed description of the problem we want the language model to solve. Given a natural language description of a system of interest, it should provide a related CRN (cf. Section 2.1) adhering to the formal grammar (given here in ISO EBNF [1]):

```
root        = "```\n" reaction, {reaction}, "```\n";
reaction    = [species], " -> ", [species], " @ ", rate, ";\n";
species     = [coeff], speciesName, [" + ", species];
coeff       = (*Number between 2 and 9 inclusive*);
speciesName = (*one letter, followed by letters, numbers or "_"*);
rate        = (*Basic floating point number, e.g., 3.14*)
              | "k", (*whole number, e.g., 0, 1, 2, ...*);
```

For example, the description

"*A chain reaction occurs from A to B over C. B decays with rate 4.2.*"

should result in the model

````
```
A -> C @ k0;
C -> B @ k1;
B -> @ 4.2;
```
````

One could argue that the description already closely adheres to the formal model. However, performing this task is still non-trivial, as it requires attending to the correct parts of the sentence to identify the involved entities and rates, as well as delineating the products and reactants (where the products might sometimes occur before the reactants and vice versa) and being robust against, e.g., spelling mistakes. Further, the interpretation of a "chain" of reactions must be known. After specialization on the above task, we require that the language model can still exploit its generality to allow for interaction with the user. For example, after the above conversation, a domain expert might decide to "*increase the rate of decay to 4.3.*" In this case, the model should respond to such requests with an updated formal description (cf. Section 5.10).

### 4.2 Data Generation

Training data is essential to prepare an LLM for a new task. While many open-source fine-tuning datasets exist, none adequately cover our specific use case. Manually curating a dataset, e.g., by describing models from model databases, would require an enormous effort to yield a high-quality (consistent, no syntactic errors, balanced across domains, ...) dataset. Notice, however, that it is straightforward to draw samples from the language defined by the grammar above. Hence, we here rely on generating synthetic data in a rule-based manner, which can also inform a decision on whether it is worth investing more time into curating real-world datasets.

*4.2.1 Assumptions.* We made some assumptions to sensibly restrict the scope of our case study. First, we target the three application domains: systems biology, ecology, and epidemiology. We generate only reactions with at most two reactants (also referred to as *binary* reactions) and up to three products. These can cover a broad spectrum of possible reactions across domains. Further, we only allow rate constants (1.2) or variables ($k_1$) and not arbitrary functions for the kinetics, effectively assuming single-parameter kinetics (e.g., mass-action) for all reactions. Also, note that our grammar focuses on the dynamics and does not yet provide means to specify initial population counts. These restrictions may be lifted in the future by expanding the grammar and rules in the data generation. Finally, as mentioned in Section 1, the goal of this dataset is to prepare the model for translation rather than to provide it with domain knowledge. Thus, the generated networks do not have to relate to real-world systems. Regardless, the jargon of the domains, e.g., "dissociation" in systems biology, needs to be correctly reflected.

*4.2.2 Ingredients of the Generation Process.* We start by listing classes of reactions that may occur in the three application domains. These include complexation, catalysis/enzymatic, and chain/cascade reactions (systems biology), as well as mating and predation reactions (ecology). Further, the concepts of degradation/death, production/birth, and "unspecified" (supporting reactions beyond the classified ones) are part of every domain.

Each of these concepts implies a certain number and/or form of reactants, products, and reactions. It is straightforward to generate grammatically correct models according to these constraints. A large assortment of possible species names, sentence templates, and additive connectives can then be used to accurately describe the generated system in a variety of ways. Species names were taken from the UniProt [12] database (systems biology, 33, e.g., "RPS16B"), animal names such as "Fox" or "Rabbit" (ecology, 15), and disease states such as "Infected" or "Vaccinated" (epidemiology, 20). In the ecology domain, we additionally use a list of ten possible attributes, such as "distressed" or "healthy" to make species names more realistic. An initial list of sentence templates was generated with ChatGPT [49]. Note that we found generating correct description-CRN pairs directly to be too unreliable. These were then manually filtered and expanded to include all reaction classes above, yielding 370 sentences. To adequately make use of them, they were classified in terms of the domain and reaction class they belong to, as well as the number of reactants and products they describe (zero, singular, plural, and any). Further, around half of the sentences (166) describe reactions without specifying a rate, e.g., "B decays." Each template includes placeholders like `{reactants}`, `{products}`, and `{rate}`. Some special placeholders like `{reactants1}` and `{reactants2}` are used to distinguish, e.g., substrate and enzymes or predator and prey. A single sentence may also describe multiple reactions, e.g., "A and B decay." describes the two reactions, $A \longrightarrow; B \longrightarrow$. On the other hand, multiple sentences may also describe a single reaction, e.g., "B decays. Its rate of decay is 0.1". We implement the grouping of reactions and relational sentences as an additional concept for decay and production reactions in all domains. The relational sentences are taken from a separate table, with a total of 50 entries, that was manually built from the sentence template table. Finally, we came up with six additive connectives, e.g., "Additionally", that can be used at the beginning of sentences.

This leaves us with five ingredients to the generation: species names/attributes, (relational) sentence templates, and additive connectives. To expand the understanding of the LLM trained on the synthetic data, these ingredients can easily be extended to include further concepts, such as phosphorylation or mRNA transcription.

*4.2.3 Generation Process.* To obtain a single description-CRN pair, we first uniformly select one of our three domains. We then choose the number of concepts in the output CRN uniformly at random between two and four, which may lead to a number of reactions between two and around ten (considering multi-reaction sentences). Similarly, we draw three to five species names from the domain-specific list. For ecology, we also choose some attributes. The domain, concept, and species list are then used to generate reaction components (reactants, products, rates, enzymes, ...) implementing the chosen concept in line with our grammar. For example, for the "mating" concept in ecology, two reactants of the same species are chosen and assigned the attributes "male" and "female". These also form the products, including an additional "pup" of this species. As another example, "complexation" in systems biology takes two species A and B and results in a single species AB (the complex). Based on the concept and domain, a sentence template (and possibly an additional relative sentence template) is chosen, which is then filled with the information provided by the reaction generator. In this process, lists of species are verbalized by introducing commas, "and", and number words in appropriate places. We also ensure that attributes are naturally integrated into a sentence, e.g., "female healthy Fox." instead of the species name "Fox_healthy_female". In the final step, additive connectives are added between sentences with a 50% chance and capitalization at the beginning of sentences is handled. Appendix A shows some example data.

### 4.3 Few-shot Prompting and System Prompts

A unique characteristic of LLMs is that they can perform many NLP tasks that previously required task-specific approaches, like sentiment analysis and translation [20]. By formulating the task in natural language and providing the input to work on, e.g., *"translate from German to English: Simulationsmodell."*, an LLM might thus already be able to deliver adequate outputs ("simulation model"). This is called *zero-shot* prompting. To prepare an instruction-tuned LLM for a specific task, it is common to also provide it with exemplary instruction-response pairs. This *few-shot prompting* is achieved by prepending the model's input sequence with a respective conversation or alternatively providing meaningful examples in the prompt, e.g., *"Change to lowercase. For example, 'All clEar.' → 'all clear.' "*. The idea is that these examples, providing the context for the next completion, steer the answer toward the right form. In our evaluation (cf. Section 5), we use examples from our training dataset and prepend the interactions to the chat history (Appendix C compares to inclusion in the first prompt).

In addition, models can also be trained to adhere to a *system prompt*. This specially marked message at the beginning of the conversation can be used to prompt for a specific tone or give some instructions for the language model to consider when answering. We use this technique to set up the role of our model:

> "*You are a translator that translates from natural language descriptions to formal reaction system simulation models. Do not generate anything except the formal output. Do not provide any explanation. Closely adhere to the provided example syntax. When mentioning entities in the formal model, try to match their names precisely to their mentions in the textual description.*"

We arrived at this specific prompt by trial and error. For example, after observing that the model often tried to provide elaborate explanations of what it had generated, we included the sentence "Do not provide any explanation."

### 4.4 Parameter-Efficient Fine-Tuning with LoRA

When the combination of system prompt and zero-shot or few-shot prompting are not enough to enable a pre-trained LLM to perform well on a given task, the model weights can be *fine-tuned* on new data, i.e., further optimized to generate samples that fit the new data's distribution as well. As even for a small to moderate number of parameters, this is a memory-intensive process, a large variety of *parameter-efficient fine-tuning* methods were developed [62]. These do not require adjusting all model parameters but only a smaller fraction or a small set of additional parameters.

A very popular method is LoRA [30]. The central idea behind this approach is to represent the *update*, applied during training to the weights of a single neuron layer in the original LLM, as two separate matrices. In a fine-tuning process including all parameters of a layer, all weights $W \in \mathbb{R}^{d \times k}$ are adjusted by some change $\Delta W$. Instead of using full fine-tuning to calculate the update $\Delta W$ (adjusting all $d \times k$ parameters), LoRA decomposes $\Delta W$ into a product of two much smaller matrices $AB$ with $A \in \mathbb{R}^{d \times r}$ and $B \in \mathbb{R}^{r \times k}$. By only adjusting the values of $A$ and $B$, the model can then be fine-tuned much more efficiently, and the new model weights $W' = W + \Delta W$ are approximated by $W' = W + AB$. This process can be applied at multiple selected layers of the LLM. The dimension $r$ is called the *rank* and determines the number of new trainable parameters. Also, accounting for the learning rate of full training, the parameter $\alpha$ is introduced as a *scaling* factor so that the update becomes $\frac{\alpha}{r}AB$. Further, typical regularization techniques like *dropout* (randomly removing certain network connections) can be used.

To fine-tune the language model to our task (Section 4.1), a LoRA can be trained on the synthetic data (Section 4.2). By choosing an appropriate rank, scaling, and dropout rate, an LLM can be specialized to provide a formal specification when prompted. Section 5.6 will evaluate which values of rank, alpha, and dropout rate lead to good performance and how the training is done in our case in detail.

### 4.5 Grammar-Constrained Decoding

Even when fine-tuned, the language model might not reliably produce syntactically valid models. This can be enforced by constraining the selection of tokens in the output sequence to conform to a specific format, e.g., a formal grammar. We here use the approach proposed in [23] and implemented in the `transformers-cfg` extension [19] to the Hugging Face `transformers` library [60]. By providing the context-free grammar shown at the beginning of this section in the extension's (G)BNF format, the output of the LLM is forced to conform with a syntactically valid CRN.

## 5 Evaluation

This section extensively evaluates the performance of the disussed methods. We pay particular attention to randomness in the training and generation, which is often overlooked in machine learning [37]: even with the same hyperparameters, the results of training a model can differ a lot, necessitating a sufficient number of replications and determining the confidence in the reported measures. Further, due to the stochastic selection of tokens during decoding, the model outputs will also vary between runs, depending on the temperature hyperparameter. For any temperature greater than 0, we thus also need to perform multiple testing replications to gain reliable insight.

We first introduce the tested LLMs, data, hardware, and testing strategy. To begin the evaluation, we perform two preliminary tests: varying the number of few-shot examples and scanning the hyperparameter space for LoRA training. These provide the basis (optimal number of examples and best performing LoRA) for our comparison. The main part then compares the accuracy of eight different combinations of the strategies discussed in Section 4 (cf. Figure 1):

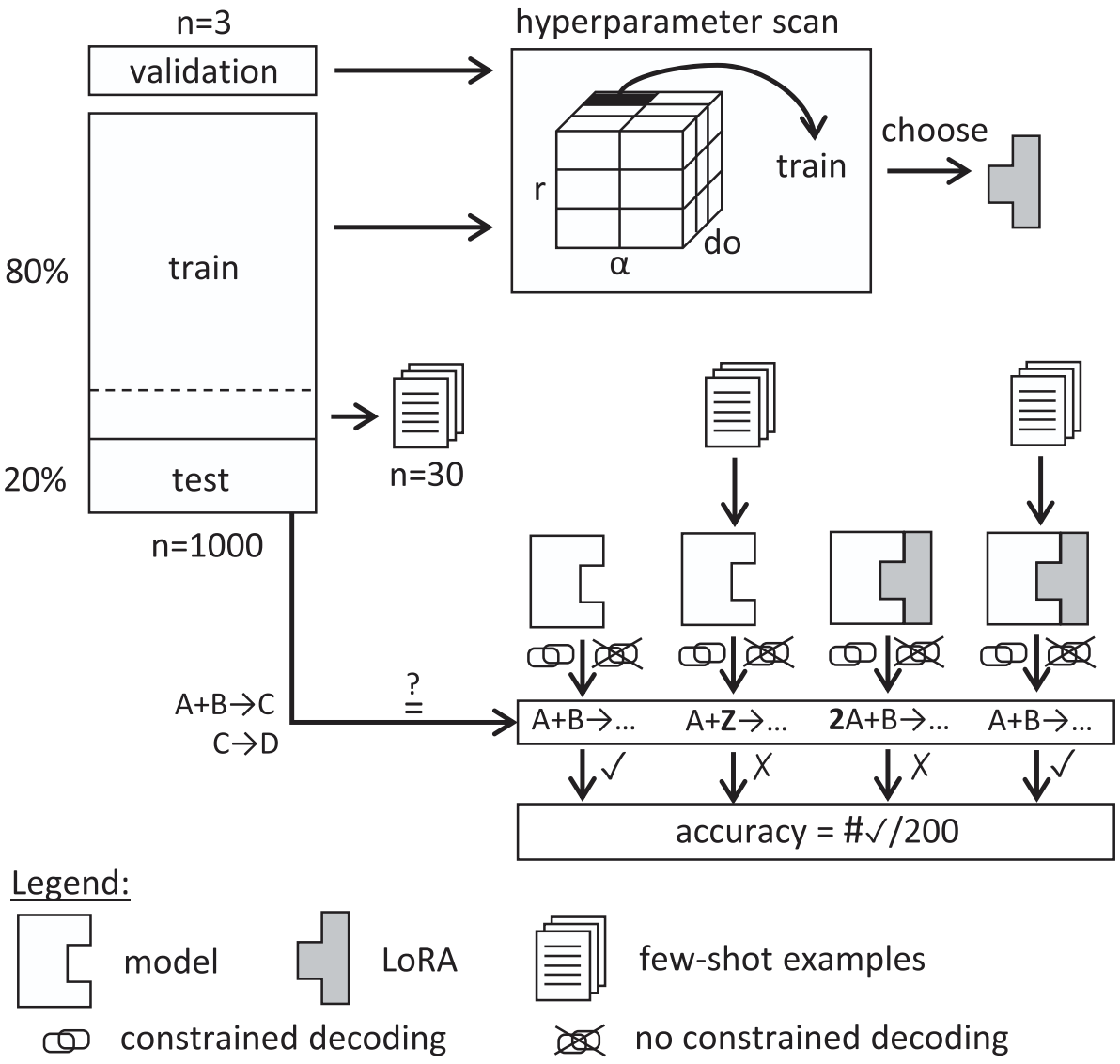

Fig. 1. Overview of experiments performed with the Mistral$^{7B}_{v0.3}$ model. We append /fs (few shot) and /LoRA to the former to refer to the different combinations.

— with or without fine-tuning a model using LoRA,
— with or without prepending few-shot examples during testing,
— with or without GCD.

We further compare these results to the few-shot performance of the larger Llama [41] and commercial GPT-4o [15] models, as well as the existing approach KinModGPT [44] and the model assembly system INDRA [27]. Finally, we conclude the evaluation by testing the practical capabilities of the best combination found in a small user study, collecting the feedback of five scientists with different amounts of modeling experience. All artifacts necessary to reproduce the results in this paper are archived on Zenodo.[1]

## 5.1 Models

The goal of our evaluation is to specifically investigate the capabilities of LLMs with a lower-end parameter count. These can be run and trained in a compute-resource-efficient and memory-resource-efficient manner and, hence, possibly on consumer hardware. Additionally, to ensure full reproducibility and freedom of use, we require at least the model weights to be available under an open-source license (open-weights). For the evaluation, we thus build on the Apache 2.0 licensed "Mistral Instruct v0.3" model with 7.3 billion parameters [33]. In the following, this model will be abbreviated as Mistral$^{7B}_{v0.3}$. As defining features, this model uses grouped query attention and sliding window attention for improved inference.

We found the "Llama instruct v3.1" model [41] with around 8 billion parameters (Llama$^{8B}_{v3.1}$) to perform similarly in our preliminary (manual) tests, but this model is only available under a restrictive license agreement and more difficult to fine-tune within our memory budget. However, we will compare its few-shot performance to Mistral$^{7B}_{v0.3}$'s on our final dataset in Sections 5.8 and 5.5. To interface with the above models and methods discussed earlier (few-shot learning, PEFT, constrained decoding), we employ the `transformers` Python module provided by Hugging Face [60].

[1]https://doi.org/10.5281/zenodo.15198397

We also compare the former to a very large, commercial LLM and test the few-shot capabilities of the "GPT-4o"[2] model by OpenAI [15, 50]. This is their flagship model at the time of writing [51]. We interface with it using the company's Python API. Note that, as opposed to using `transformers`, this may involve preprocessing (e.g., safety filters) out of our control.

### 5.2 Data

We use our synthetic data generation tool, introduced in Section 4.2, to generate instruction-response pairs, building a corpus of 800 pairs for training and 200 pairs for testing purposes (cf. Figure 1, left). To avoid a model performing well when overfitting the training data, we base the training and testing samples on disjoint sets of ingredients (e.g., species names). To this end, we group sentences by their domain, concept, and possibly other attributes, and split each group into two parts, one containing 80% of the templates (for training data generation), and one containing 20% of the templates (for testing data generation).

A third dataset is used for validation during LoRA training (cf. Figure 1, top left). This set is used to make an informed choice on when to stop the gradient descent training, a technique commonly called early stopping. The validation data consists of three manually written samples. In contrast to the training samples, these include not only a single interaction to translate to a formal model but an additional request by the (fictional) user ("Change the rate of…", "I was mistaken…") and an expected response. We hope that stopping the training when the accuracy on these samples decreases, while the accuracy increases on the training samples, contributes to the LLM retaining its original interaction capabilities and avoids overfitting.

### 5.3 Hardware

The evaluation was split across three machines. The first two are each equipped with a single NVIDIA RTX A5000 **graphics processing unit** (**GPU**) with 24,564 MiB[3] of graphics memory (VRAM). The third has an architecturally very similar NVIDIA RTX 3090Ti GPU with the same amount of VRAM. When using and training LLMs, the graphics memory is the limiting factor, determining how large a batch (subsample of the data used with stochastic gradient descent training [42]) can be and how many parameters can be added by a LoRA. We chose a batch size of 10, which is the largest that still fits in the graphics memory, and used a gradient accumulation of 80 to emulate an effective batch size of 800, making all training samples fit into a single batch. We also make use of the bf16 floating-point representation for training and load the locally running models in the 16-bit floating-point format.

### 5.4 Testing Methodology

To evaluate performance, we sequentially prompt the LLM with the instructions from our testing dataset and determine its accuracy, which is defined as the number of (semantically) correct answers out of the possible 200 (cf. Figure 1, bottom right). E.g, if an LLM would produce 150 correct outputs (and, consequently, 50 incorrect ones), its accuracy would be determined as 75%.

*5.4.1 Equivalence of Models.* In our case, correct means aligning with the ground truth answer from our test data. We test this alignment by extracting the reactions from the model output and checking whether, for each reaction in the ground truth, there is a corresponding reaction in the model's answer.

Two reactions are corresponding if they share the same reactants, products, and rate constant. To test the equivalence of reactants and products, we transform both the ground truth species names and the species names extracted from the model's output to lowercase and disregard their order. We do convert to lowercase characters, as only the consistency matters for the formal output, i.e., the same species names have to be used in the right places without regard to their capitalization. For example, if a user uses different versions of the same species name ("Wolf" and "wolf"), the LLM should be free to decide on a single version. This also circumvents some problems with capitalization at the beginning of sentences, where the spelling of the species name is ambiguous. Besides simple equality of the names in lowercase, we also allow different orders for attributes (e.g., "male_wolf" instead of "wolf_male"). Finally, we also check for the correct coefficient (e.g., the 2 in "2wolf"). For the rate constant, there are two possible cases: the rate is specified as a floating point number, in which case they have to match exactly, or it is defined as a variable, in which case we only check if it starts with the letter "k", as required by our grammar.

[2] At the time of writing pointing to the version gpt-4o-2024-08-06.
[3] Value reported by the `nvidia-smi` tool.

*5.4.2 Stochastic Test.* The above method yields a point estimate of the accuracy. However, when considering a temperature parameter greater than zero, the outputs of an LLM are stochastic. To evaluate the performance in this setting, we follow a stochastic convergence test according to [29]. This procedure performs replications of a test (here, a single walk through all 200 testing examples) until the mean of the reported measure (here, the accuracy) converges within certain bounds. We require the half-width of the confidence interval with confidence level 0.99 to be $d_n = 0.02$, i.e., 2% and report the converged mean estimate and the corresponding standard deviation. At least $n = 3$ initial replications are performed for the initial estimate, and we require $kLimit = 2$ subsequent iterations to be within the bound $d_n$ before declaring convergence.

### 5.5 Number of Few-Shot Examples

As outlined in Section 4.3, LLMs may learn by example to complete certain tasks. We start our evaluation by testing the accuracy of all three discussed LLMs when varying the number of such examples. For this preliminary test, we set the temperature to 0 to obtain deterministic results and avoid time-consuming stochastic replications.

We incorporate few-shot examples by prepending prompt-answer pairs from the training dataset in the context window of the LLM according to its respective chat format (prepend them to the "chat-history"). An alternative would be to include examples in the first prompt, which we found resulted in a similar but slightly worse performance (cf. Appendix C). The tested values are the first 0 (zero-shot), 1, 5, 10, 20, 30, 40, 50, 60, and 70 examples from the training dataset. Where possible, this includes at least one example from each domain in the conversation. This corresponds to 75 (zero examples, just system prompt) to 9,279 (70 examples) tokens, when using the tokenizer of the $\text{Mistral}^{7B}_{v0.3}$ model (93 to 8,526 for $\text{Llama}^{8B}_{v3.1}$). Note that the Mistral model's context length is just 8192 tokens [33], which is exceeded after around 60 training examples when also including a prompt.

The results are summarized in Figure 2 and show that an increase in the number of examples generally leads to an increase in accuracy. However, for any number above 20 to 30 examples included, there are diminishing returns. This saturation of the accuracy can be explained by the model's intrinsic few-shot capabilities being exhausted and also the way we include the examples sequentially from our (randomly generated) training data. For lower numbers of examples, not every possible construct (e.g., chain reactions; see Section 4.2) is included, which leaves the model more freedom to deviate from a correct translation. With more examples, the likelihood of every construct being inside the context window of the LLM increases. It can also be seen that (at least at temperature 0), $\text{Mistral}^{7B}_{v0.3}$ only arrives at around half of the accuracy achieved by the 1B parameter larger $\text{Llama}^{8B}_{v3.1}$. The $\text{GPT}_{4o}$ model achieves a very high accuracy of 86% with only five examples. Analyzing the details, the accuracy in ecology and epidemiology is 95% at this point vs. 68% in the systems biology domain. This is likely the case, as only a single example from the systems biology domain is included in the first five. With a sufficient number of examples, however, GPT$_{4o}$ achieves a 99% accuracy, demonstrating its superior generalization capability over the smaller models. From this preliminary experiment, we deem a number of 20 (Llama$^{8B}_{v3.1}$) and 30 (Mistral$^{7B}_{v0.3}$ and GPT$_{4o}$) few-shot examples as optimal for the subsequent tests.

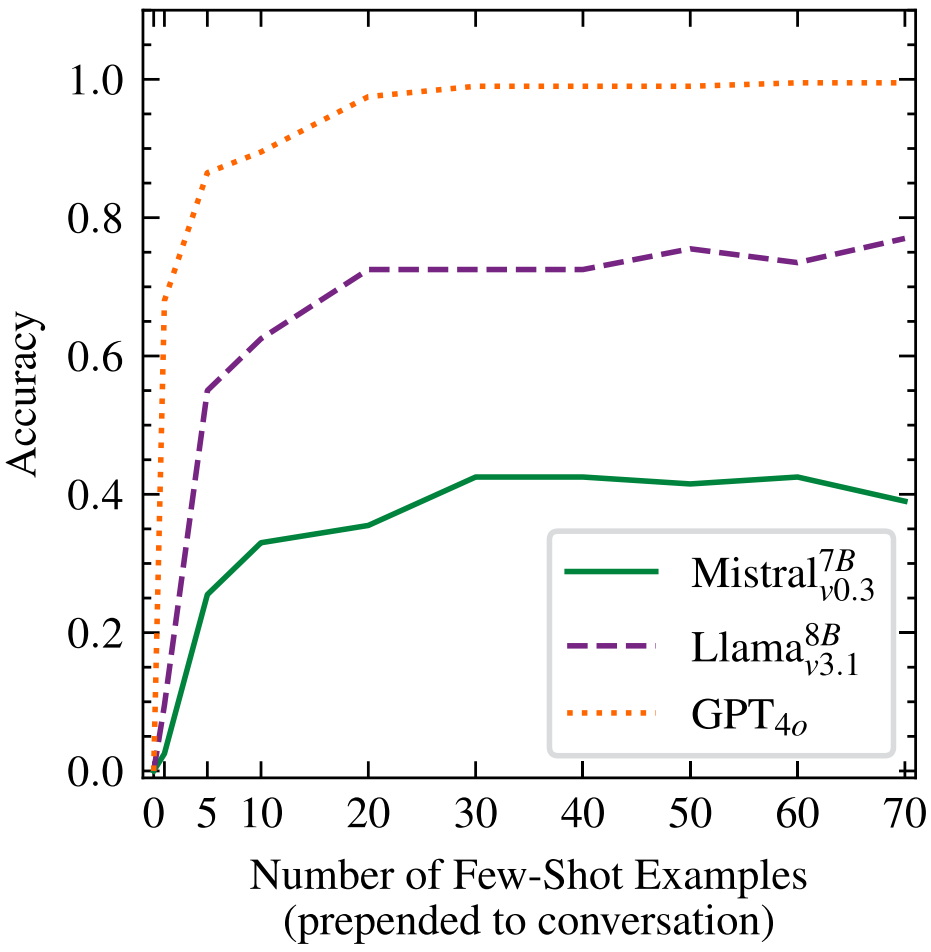

Fig. 2. Performance at temperature 0 when varying the number of few-shot examples included in the conversation for the Mistral$^{7B}_{v0.3}$, Llama$^{8B}_{v3.1}$, and GPT$_{4o}$ models. Note how there is no significant change beyond 20 to 30 examples.

## 5.6 LoRA Hyperparameters

Besides relying solely on the few-shot capability of a language model, it is possible to adjust the weights that define it. This can be done in a parameter-efficient and, thus, memory-efficient way using LoRA (cf. Section 4.4). In our second preliminary experiment, we scan important hyperparameters of the LoRA training with Mistral$^{7B}_{v0.3}$ to increase the chance of obtaining the best possible model (cf. Figure 1, top right). This is important, as these may have a large influence on the accuracy. As the theoretically most important hyperparameters for LoRA, we chose the rank $r$ and scaling factor $\alpha$. The former determines the number of trainable parameters added to the model, while $\alpha/r$ scales their influence on the original weights. Also, the *dropout (do)* percentage seemed to have a significant impact on preliminary tests. This is a regularization technique that randomly excludes weights of a neural network during training, forcing them to generalize instead of overfitting.

We train a LoRA (cf. Section 4.4) for all linear layers of the Mistral$^{7B}_{v0.3}$ model on our training dataset to see how this impacts its performance on our test dataset. To evaluate the influence of hyperparameters in our case, we perform a grid search over 18 combinations of rank $r \in \{4, 8, 16\}$, $\alpha \in \{8, 16\}$, and $do \in \{0.3, 0.5, 0.7\}$. This results in covering a wide range of scaling factors between $\alpha/r = 8/16 = 0.5$ and $16/4 = 4$ (cf. Appendix B). The number of new parameters introduced are $10,633,216$ / $21,266,432$ / $42,532,864$ in the case of $r = 4, 8, 16$, respectively. Even when using the same hyperparameters, there may be differences in accuracy due to the stochastic training procedure and initialization. Hence, we repeat the complete grid search three times with a different seed. Note that three samples are insufficient to get an accurate estimate of the mean (and standard deviation) but still provide good enough grounds to make an informed choice. As optimizer, we employ the default AdamW [42] optimizer and hyperparameter settings provided by the Hugging

Face transformers API. We set the learning rate to $10^{-4}$, as we found the optimization to converge faster than with the default of $10^{-5}$ without quality loss. Further, we chose a batch size of 10 and gradient accumulation of 80 (cf. Section 5.3).

The grid search with three replications results in $18 \times 3$ LoRAs, which need to be tested. As in our previous preliminary experiment, we do so at temperature 0 for deterministic results. This enables us to elude the very time-consuming stochastic tests, so we can do a single evaluation for each of the LoRAs to assess their quality. For each of the 18 LoRAs, we evaluate its configuration by the mean accuracy and its standard deviation.

The results of the hyperparameter grid search are summarized in Table 2 in Appendix B. It is evident from the results that there is no systematic way in which the hyperparameters influenced the accuracy. All combinations result in an accuracy between 70% and 80%. This is likely explained by the task being solvable with only a few additional trainable LoRA parameters. We hypothesize that the hyperparameters may play a much more significant role when using more complex DSLs and a larger assortment of concepts in the data. The combination ($\alpha = 8, r = 8, do = 0.3$) achieves the highest accuracy. For the following evaluation, we thus choose the best-performing replicate out of the 3 LoRAs trained on this combination, that achieved an accuracy of 85% (which is also generally the highest value achieved). We refer to this model-LoRA combination as $\text{Mistral}^{7B}_{v0.3}$/LoRA.

*5.6.1 Early Stopping.* Even with relatively small values for rank and alpha, we found that using a dataset with a similar form to our training data, although having zero overlap with the training data, resulted in overfitting. Whereas this does not mean overfitting the training samples themselves, it eliminates the model's ability to perform some of the other functions it was previously trained on, which is also known as "catastrophic interference" [47]. To mitigate this problem, we employ a specialized validation dataset as described in Section 5.2. We found this technique has little to no impact on accuracy while increasing the LLM's recall of its previous abilities.

### 5.7 Comparing the Methods

After determining an appropriate number of few-shot examples and a well-performing LoRA, we can compare the performance of the different combinations of methods on $\text{Mistral}^{7B}_{v0.3}$ as summarized in Figure 1 (bottom right). Our basis is the bare model and model with added LoRA. Each of them can be augmented by providing 30 (as determined earlier) few-shot examples. In turn, each of these four combinations can be combined with constrained decoding (cf. Section 4.5). This leaves us with eight combinations to test. Note that we still use a single example for testing the bare model, as otherwise the model has no indication of the expected output format. However, when testing $\text{Mistral}^{7B}_{v0.3}$/LoRA, we actually test zero-shot performance.

To get a better overview of how the generation performance varies across temperatures, this time, we also scan over temperature values $t \in \{0.0, 0.2, 0.4, 0.6, 0.8, 1.0\}$. The results, summarized in Figure 3, show that without using GCD, at a temperature of 0.2, the few-shot approach achieves its best performance of $43.9\% \pm 1.8\%$. As expected, the accuracy decreases with increasing temperature as the model is given more freedom in choosing (improbable) next tokens. On the other hand, the fine-tuning approach achieves its maximum accuracy of $84.5\% \pm 0$ at temperature 0.0. Thus, by using a LoRA trained on synthetic data, $\text{Mistral}^{7B}_{v0.3}$ can achieve an increase in accuracy of 41% compared to few-shot prompting. This showcases the power of (parameter-efficient) fine-tuning for small models as opposed to just using a few-shot approach, as predominantly used in related work (cf. Section 3). It also shows that even locally run LLMs with relatively few parameters may achieve passable performance.

When using constrained decoding, the results are similar, but often this technique seems to slightly decrease the accuracy in our case. We attribute this to the fact that, already with the fine-tuning or LoRA, the LLM learns to accurately adhere to the grammar as implicitly defined by the training data. Adding additional restrictions to decoding seems to mostly introduce additional (semantic) errors, e.g., swapping reactants and products. However, just as discussed for the LoRA hyperparameters, this result might be different in cases where the grammar is more complex and not easily picked up from examples. Also, our user-study (Section 5.10) revealed rare cases where there was a syntactic error when only using LoRA.

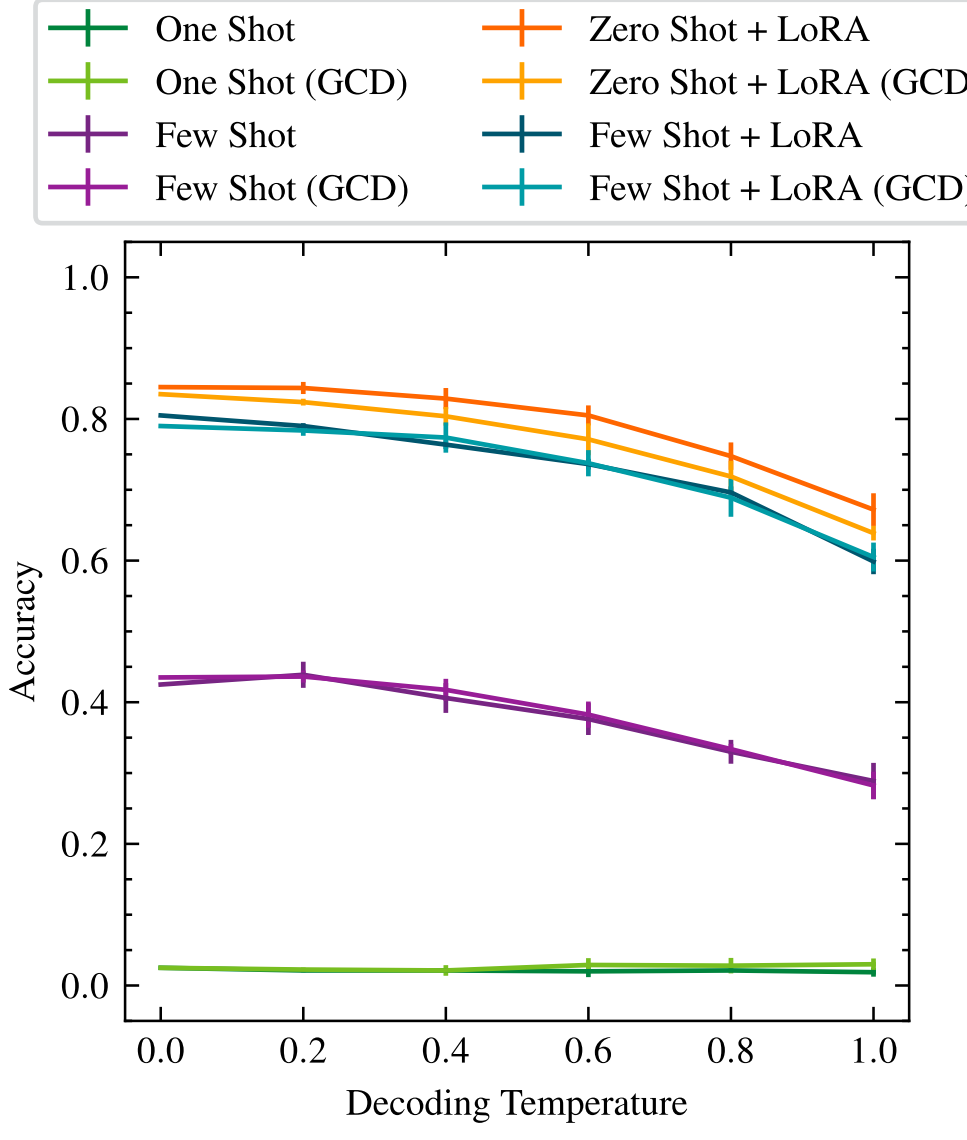

Fig. 3. Comparison of accuracy at different temperatures when combining few-shot prompting, LoRA, and GCD on Mistral$^{7B}_{v0.3}$. The error bars visualize the standard deviation over different LLM decoding seeds.

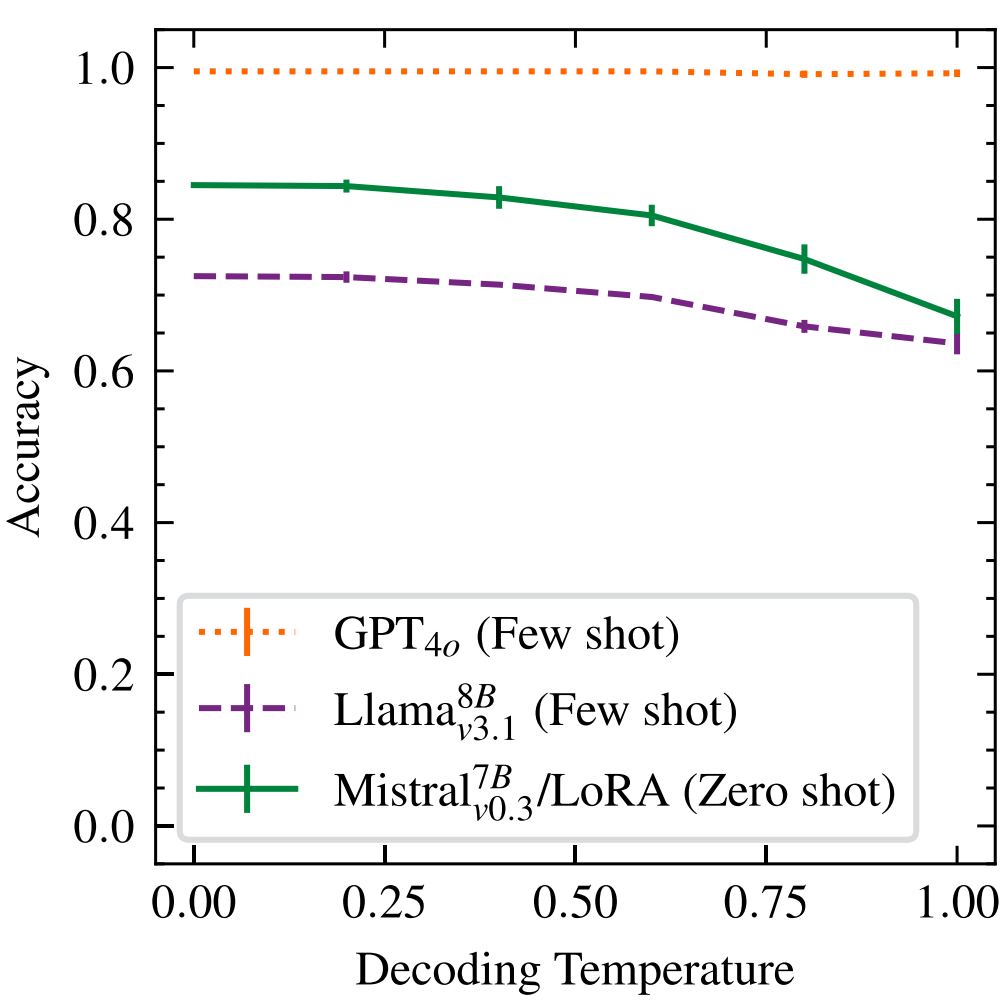

Fig. 4. Few-shot performance of GPT$_{4o}$ and Llama$^{8B}_{v3.1}$ for different temperatures compared to Mistral$^{7B}_{v0.3}$/LoRA.

### 5.8 Comparison to Llama 8B and ChatGPT

We showed how Mistral$^{7B}_{v0.3}$, an LLM with relatively few parameters, can be adapted to perform well in generating simulation models from textual descriptions. We found the accuracy to be promising, although not yet production-ready. However, the approach we presented can be used with a very low bar of entry regarding the required hardware and using an open-weights model. It is thus interesting to see whether a larger, commercial LLM can use its superior generalization capability to perform this task just by few-shot prompting.

To this end, Section 5.5 already showed that, with the right number of examples, GPT$_{4o}$ can outperform Mistral$^{7B}_{v0.3}$ when using few-shot prompting. To put this into perspective with Mistral$^{7B}_{v0.3}$/LoRA, Figure 4 shows the results of repeating the temperature scan from the last section with 30 examples for GPT$_{4o}$ and 20 for Llama$^{8B}_{v3.1}$. It is evident, that the orders-of-magnitude larger[4] LLM can still outperform the specialized, smaller LLM by just providing examples. However, the 1B larger Llama$^{8B}_{v3.1}$, with few-shot prompting, can not outperform the specialized Mistral$^{7B}_{v0.3}$. Note also that model size seems to increase robustness at higher temperatures. We expect that developments in the currently very active small and open-source LLM space [18] and improvements in the training data may close the gap to larger models in the future. Fine-tuning slightly larger (than 7B) models is also an important direction for future work, as indicated by the results achieved with $\text{Llama}^{8B}_{v3.1}$. Finally, Table 1 summarizes the accuracies achieved by the best models we evaluated.

[4]No public information is available. Estimates range from $10^{12}$ to $10^{14}$ [39], which is between three and five orders of magnitude more than Mistral$^{7B}_{v0.3}$.

Table 1. Best Accuracies Achieved Across Tested Models and Methods (LoRA Fine-tuning and Few-shot Prompting)

| **Model** | $\text{Mistral}^{7B}_{v0.3}$/LoRA | $\text{Mistral}^{7B}_{v0.3}$/fs | $\text{GPT}_{4o}$/fs | $\text{Llama}^{8B}_{v3.1}$/fs |
|---|---|---|---|---|
| **Accuracy** | 84.5% ± 0% | 43.9% ± 1.8% | 99% ± 0% | 72.5% ± 0% |

### 5.9 Comparison to other Tools

In the following, we compare $\text{Mistral}^{7B}_{v0.3}$/LoRA to KinModGPT [44] (paired with $\text{GPT}_{4o}$), as well as the model assembly engine INDRA [27]. Both tools are designed for the systems biology domain, so we restricted our dataset to the 62 biochemical examples for the following tests.

The comparison to KinModGPT can only be made approximately, as some features/domains are only supported by one of the tools. For example, KinModGPT allows defining initial concentrations and uses rate functions to translate, e.g., mRNA translation, which our grammar does not yet support.

To start the comparison, we evaluated the accuracy of the $\text{Mistral}^{7B}_{v0.3}$/LoRA on the small benchmark set of four example models “Decay” (1 reaction), “HIV” (10 reactions), “Three-step” (15 reactions), “Heat shock response” (50 reactions) used in [44]. We found that taking the (current) assumptions made in our training into account, we could achieve results similar to KinModGPT (cf. Appendix E). When disregarding the initial concentration in “Decay”, which $\text{Mistral}^{7B}_{v0.3}$/LoRA expectedly interpreted as a reaction, the “Decay” and “HIV” models could be translated without error. For the “Three-step” model, while the chain reaction at the beginning is translated correctly, our model was not able to relate to the first reactions as “metabolic” and incorporate the enzymes, trying to construct separate enzymatic reactions. Further, some reactions require the use of rate functions. As we only allow constant rates, these reactions cannot be translated properly. The output is still able to approximate these constructs, just ignoring the rate. In the very large heat-shock response model, the first half of 25 reactions is translated correctly, except for three missing reactions that $\text{Mistral}^{7B}_{v0.3}$/LoRA unexpectedly ignored. The second half, which, similar to “Three-step”, requires many features not yet supported, translates the supported reactions correctly, i.e., the unknown constructs don’t seem to impact the performance. This shows that $\text{Mistral}^{7B}_{v0.3}$/LoRA is also able to handle natural descriptions far beyond the maximum length it was trained on.

To complete the comparison, we also adapted the KinModGPT approach to work with our formulations and output format. As it relies solely on engineering a specific prompt for large LLMs, we achieved this by adjusting their system prompt as shown in Appendix D. We performed a test run as described in Section 5.4 using $\text{GPT}_{4o}$ at temperature 0, which resulted in a 95.2% accuracy. This provides further evidence for the applicability of KinModGPT beyond the four examples tested in [44]. Taking into account the previous findings, where $\text{GPT}_{4o}$ achieved up to 99%, adding more example interactions to the prompt might further increase the accuracy.

Finally, we compare $\text{Mistral}^{7B}_{v0.3}$/LoRA to the INDRA tool, which uses traditional reading engines. In our case, we used the TRIPS [3] processor to test our 62 biological examples, following the INDRA tutorial and the implementation in [44]. In only around 42% of cases did the tool yield a non-empty model, with none of them completely achieving the desired outcome. Almost always, only a small part of the reactions were identified, often with a wrong rate constant. We noticed that it seemed to only produce an output if it found the species contained in (and identified from) the description to be part of one of its connected databases. However, even though all species names originated from the UniProt database [12], it failed to identify many of them and produce models. We attribute this mainly to the fact that INDRA was developed with a slightly different purpose in mind than we pursue here. Their target descriptions ("word models") follow very specific patterns and phrasing. Hence, whereas INDRA excels at building pipelines for word model processing, filtering, and manipulation, an LLM may understand more natural/ambiguous formulations, but lacks the "mechanistic pipeline" parts of INDRA. An interesting future research direction is thus the development of a transformer-based reading engine [35].

### 5.10 Small-scale User Study

We concluded the evaluation with a small-scale user study by interviewing $n = 5$ scientists with varying experience in developing simulation models. The following synthesizes the most important findings. A detailed summary of answers is provided in the supplemental material (S),[5] which we refer to in the following. Our goal was to reveal first insights into $\text{Mistral}^{7B}_{v0.3}$/LoRA's usability for modeling and to test its generalization capabilities beyond the synthetic test cases. All participants were at least vaguely familiar with the formalism of CRNs, and could thus rely on their experience to assess the produced formal model. In preparation for the interview, they were given information about the assistant's current limitations (Section 4.2.1) and asked to prepare descriptions of two systems of interest (S.1). Each supervised session lasted about an hour.

Participants were asked to fill out a short questionnaire about their expectations at the beginning and the end of the interview. Most participants were optimistic that the prototype chatbot would be helpful, particularly for building simple models (S.2.3). The possibility of building more complex models using natural language was met with more skepticism. Next, each participant was asked to describe a small model and then a more complex one from their area of interest by interacting with the LLM assistant. These included ecology (5), systems biology (2), epidemiology (2), and a multi-agent model (1). While performing the tasks, the participants were asked to answer a few questions about their experiences and feelings. As expected, the translation worked better for the simpler descriptions (S.3.2). More complex descriptions lead to reasonable models, but never without mistakes (S.3.7). When the first result was not as expected, the participant was asked to try to fix it by interaction. In many cases, this failed (S.3.3, S.3.8). In training the LoRA, we did not explicitly include interaction, but hoped the model would retain its interactive capabilities and would generalize. It seems that the latter is not the case, and/or training the LoRA led to catastrophic interference (Section 5.6.1). This indicates the necessity to take interaction into account explicitly in the training data. Hence, resubmitting a full (adjusted) description was typically more effective. Although this often worked (S.3.3), it also led to hallucinations in a complex model (S.3.8). After the small examples, most enjoyed the session and felt they had learned something about the assistant's behavior. This likely also slightly influenced (S.3.7) the second interaction, with more complex descriptions. However, most participants already committed to their descriptions beforehand (S.1). More complex modeling tasks proved challenging, particularly as interactive corrections were not easy (S.3.8). The assessment of the participants shifted to the negative side.

Generally, the assistant was quite robust against typos, even in species names, concluding the correct name from the context (S.3.11). This might also be an advantage over traditional reading engines (cf. Section 5.9).

The way of interaction varied. Especially participants with a lot of modeling experience stayed very close to the reaction formalism in their natural language expression, which allowed for higher success rates (S.3.3). Some participants used more general domain concepts in their descriptions. Depending on how ambiguities were resolved, the results aligned more or less with expectations.

[5] Attached to the online version of this article.

For example, one participant tried to describe a predator-prey model according to Lotka and Volterra [43, 57], that merges predator hunting and growth into a single reaction. This is however against the way we trained the model to resolve the statement "predator hunts prey" and the assistant seemed "confused" (S.3.3). We conclude that future improvements should allow the specification of preferences (also identified in [9]) or learning from past interactions. Also, some desired concepts were missing (S.4.3). This became particularly evident in the domain of epidemiology, where the basic SIR reactions ($S + I \longrightarrow 2I; I \longrightarrow R;$) could not be derived from a natural language description (S.3.2). Considering implicit formulations of "infection" not being part of training yet, this shows each such concept has to be explicitly included, even though the LLM likely encountered it in its initial training.

In the final questions, participants were asked about the suitability of the assistant as a tool for modeling and possible required improvements. They agreed that, in its current form, it may only be useful to support beginners, probably because their expectations as experts were not met. We note, however, that this might be problematic if the beginner cannot properly assess the correctness of the result. An on-the-fly execution of models may help in this regard, as done in [9]. The participants' wish list includes increased accuracy, language features (attributes, guards), improved understanding of domain concepts (activation, trimeric complexes), and straightforward interactive editing (S.4.3).

## 6 Discussion and Conclusion

In this article, our goal was to evaluate whether small, open-weight LLMs could be used to translate natural language into formal simulation models. To this end, using the seven-billion-parameter Mistral Instruct v0.3 ($\text{Mistral}^{7B}_{v0.3}$) model as example, we showed how to fine-tune (small) LLMs to translate from natural language descriptions of population-based reaction models to the respective CRN in a formal domain-specific modeling language. We extensively evaluated the accuracy achievable with synthetic data and compared the performance to the large commercial GPT-4o ($\text{GPT}_{4o}$) model and the restrictively licensed Llama Instruct 8B v3.1 ($\text{Llama}^{8B}_{v3.1}$) model. Further, we compared the fine-tuned $\text{Mistral}^{7B}_{v0.3}$ to two other methods, KinModGPT [44] and INDRA [27], which rely on a large-scale, commercial LLM and a traditional reading engine, respectively.

Our results show that $\text{Mistral}^{7B}_{v0.3}$, even with parameter-efficient fine-tuning, cannot reach the accuracy of the much larger $\text{GPT}_{4o}$(84.5% vs. 99%). However, considering the orders-of-magnitude difference in parameter count, it comes remarkably close. We think, that future developments of small and open-source LLMs, like DeepSeek's recent distill models [18], and higher-quality data can close this gap. Considering the few-shot results of $\text{Llama}^{8B}_{v3.1}$, models slightly larger than 7B may also offer improvements. Compared to INDRA, such systems lack sufficient grounding in knowledge. This might be solved by incorporating RAG into the LLM inference [46]. On the other hand, even small LLMs seem robust against noise in spelling and grammar (cf. Section 5.10).

As data for training translation to simulation models (here, CRNs) is scarce, we generated a synthetic dataset. Its generation can be improved in a variety of ways: by including more domain concepts and using ontologies to ground samples in real-world relations (Which words mean infected? Who can be predator?); using model generation tools, such as [52, 61], to generate realistic models; including attributes for all domains to represent, e.g., phosphorylation; and including kinetic functions and initial conditions. Most importantly, as evident from our user study, interaction needs to be included. This may be achieved by defining interaction concepts (e.g., rate change, reaction removal) and additional templates for interaction requests. Beyond the above, supporting higher-level concepts ("make the infection more lethal") would be interesting. Based on our user study, we believe an iterative improvement process involving users to yield the best results.

As shown in [9], LLMs can support learning to model in a specific formal paradigm. We here employed a smaller, self-hostable LLM, making it accessible to a broad audience. While our current prototype is not intended as a learning system, we believe that building on the demonstrated approach could also support modellers in learning. The participants in our user study suggested this as well. However, this requires the learner to critically assess the answers. As the model is defined in a formal DSL, it can be checked for syntactic correctness and executed, so they might be able to determine whether the generated simulation model reflects their intentions by inspecting the simulation output. The difference between the generated model's and the expected output might also be an alternative metric of accuracy. We have seen that the simple DSL we defined is relatively easy to train for and GCD is not beneficial. Still, the latter could become more relevant for more expressive modeling languages, which is an interesting direction for future developments.

Another interesting direction is to integrate the natural language approach with the automatic generation of (CRN) simulation models from data [32, 38, 45]. On the one hand, the formalized version of the natural language description could be used as an additional information source. On the other hand, it would be interesting to derive a model from a description of the model's *output data* in natural language. Thus, our findings also contribute to efforts to further automate simulation studies [55].

To conclude, we believe, based on our findings, that the recent developments in NLP with LLMs will enable entirely new opportunities for modeling by interacting with computer assistants that might even be executable on personal computers.

## Acknowledgments

We thank the anonymous referees for their valuable comments, which lead to great improvements over the initial draft; Bjarne Hiller for helpful feedback on the first draft and TikZ; Philipp Henning and Maria Pierce for helping with the jargon of biochemistry and ecology; and the five participants of our small-scale user study.

# Appendices

## A Excerpts from the Training Data

The following are excerpts of description-CRN pairs, taken from the training data (cf. Section 5.2), showcasing some of the concepts currently supported by the synthetic data generation:

"*A PDC1 molecule is transformed to GPM1 via an enzymatic reaction, catalyzed by PGK1.*":

```
PDC1 + PGK1 -> GPM1 + PGK1 @ k0;
```

"*Furthermore, a cascade of reactions occurs from HYP2 to RPL27A over RPP2A, Drp1, and cATP*":

```
HYP2 -> RPP2A @ k0; RPP2A -> Drp1 @ k1; Drp1 -> cATP @ k2; cATP -> RPL27A @ k3;
```

"*Male poisoned Rat and poisoned female Rat mate.*":

```
Rat_poisoned_male + Rat_poisoned_female -> Rat_poisoned_pup + Rat_poisoned_female +
Rat_poisoned_male @ k0;
```

"*Additionally, Fox feed on Rat.*":

```
Rat + Fox -> Fox @ k0;
```

"*Sick emerge. Their birth rate is 2.23.*":

```
-> Sick @ 2.23;
```

"*A Contagious individual progresses to the Healthy state at a rate of 0.74.*":
```
Contagious -> Healthy @ 0.74;
```

## B LoRA Hyperparameter Scan

See Table 2.

Table 2. Grid Search Over LoRA Hyperparameters, Varying Alpha $\alpha$, Rank $r$, and Dropout (do)

| alpha | rank | scale | dropout | mean | stddev |
|---|---|---|---|---|---|
| 8 | 4 | 2 | 0.3 | 0.73 | 0.03 |
| 8 | 4 | 2 | 0.5 | 0.75 | 0.03 |
| 8 | 4 | 2 | 0.7 | 0.69 | 0.03 |
| **8** | **8** | **1** | **0.3** | **0.79** | **0.06** |
| 8 | 8 | 1 | 0.5 | 0.73 | 0.05 |
| 8 | 8 | 1 | 0.7 | 0.71 | 0.04 |
| 8 | 16 | 0.5 | 0.3 | 0.77 | 0.03 |
| 8 | 16 | 0.5 | 0.5 | 0.76 | 0.04 |
| 8 | 16 | 0.5 | 0.7 | 0.74 | 0.05 |
| 16 | 4 | 4 | 0.3 | 0.72 | 0.02 |
| 16 | 4 | 4 | 0.5 | 0.74 | 0.01 |
| 16 | 4 | 4 | 0.7 | 0.74 | 0.03 |
| 16 | 8 | 2 | 0.3 | 0.73 | 0.04 |
| 16 | 8 | 2 | 0.5 | 0.74 | 0.04 |
| 16 | 8 | 2 | 0.7 | 0.76 | 0.01 |
| 16 | 16 | 1 | 0.3 | 0.74 | 0.01 |
| 16 | 16 | 1 | 0.5 | 0.74 | 0.02 |
| 16 | 16 | 1 | 0.7 | 0.74 | 0.05 |

For each configuration, three LoRAs are trained with different seeds. The mean accuracy (cf. Section 5.4) and its standard deviation are reported when testing at temperature 0. The bolded row marks the best-performing configuration, from which we selected the second trained LoRA that achieved an accuracy of 85% for further testing.

## C Few-shot Strategy Comparison

See Figure 5.

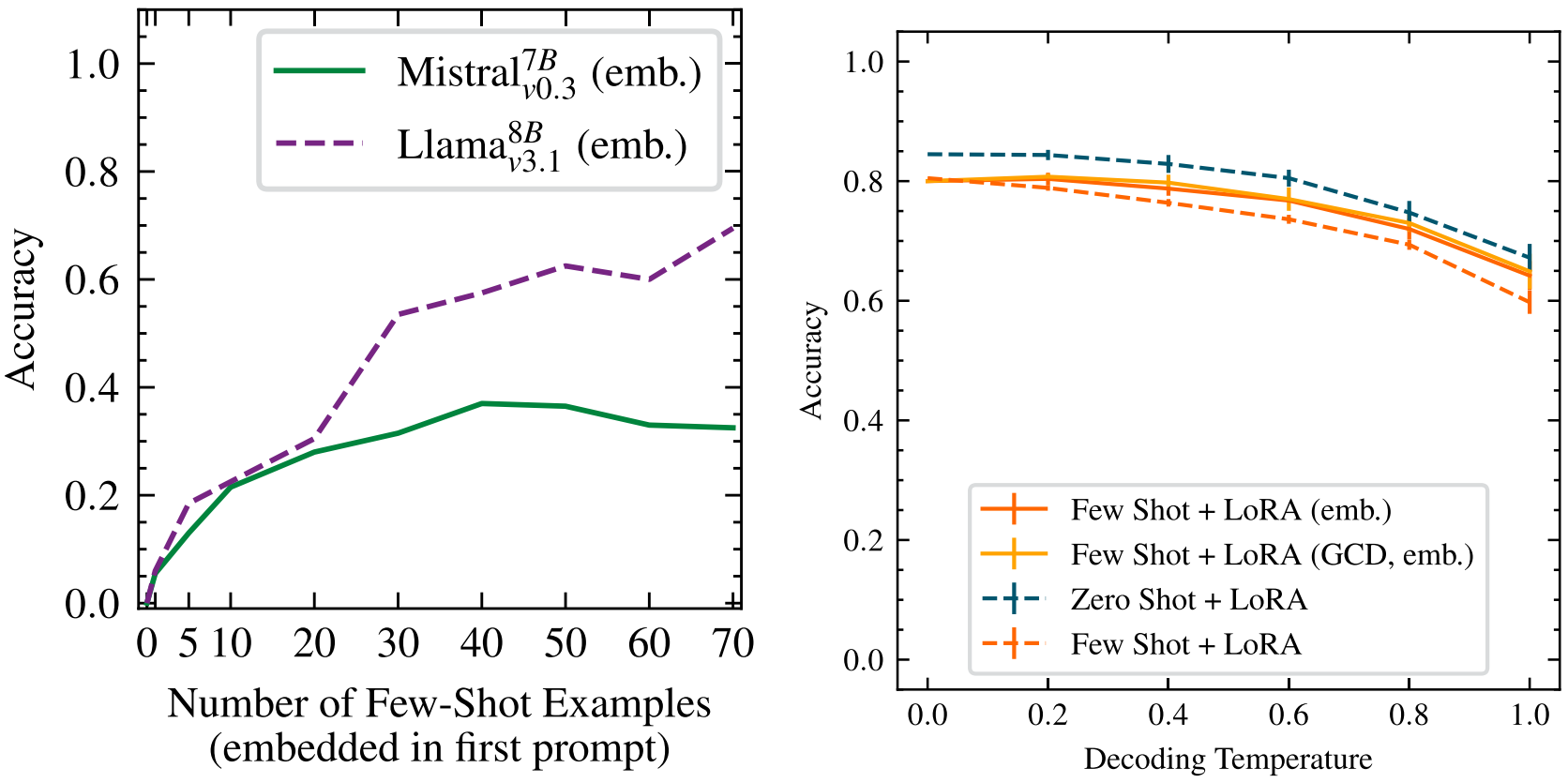

Fig. 5. (Few shot) accuracy when embedding examples.

## D Adjusted KinModGPT System Prompt

The KinModGPT approach relies solely on formulating a system prompt of conversion rules to produce statements in the Antimony language. To allow for a fair comparison, we adjusted their system prompt to produce statements according to our grammar and to work well with the features required by our dataset, as follows:

````
You are a program that converts biochemical reactions written in ↪
 natural language into a formal reaction language. First, remember the↪
  following conversion rules.

# Conversion rules
| Natural language | Antimony language |
| E catalyzes the conversion of X to Y | X + E -> Y + E @ k0; |
| X is converted into Y | X -> Y @ k0; |
| X and Y bind to form Z | X + Y -> Z @ k0; |
| X dissociates into Y and Z | X -> Y + Z @ k0; |
| X is produced (or transcribed) | -> X @ k0; |
| X degrades (or decays) | X -> @ k0; |

# Examples
"The following describes a reaction system. Please translate to a ↪
 formal description. RPL35A is produced. It is produced with a rate of↪
  2.42. In addition, GPM1 and RPL35A are removed from the system. ↪
 RPL35A emerges at a rate of 7.9." is converted to ```
-> RPL35A @ 2.42;
GPM1 ->  @ k0;
RPL35A ->  @ k1;
-> RPL35A @ 7.9;
```
"The following describes a reaction system. Please translate to a ↪
 formal description. HSP26 vanishes. It leaves the system at a rate of↪
  9.82. Two ATP are the result of a conversion of TDH3 and TPI1. A ↪
 chain reaction occurs from TDH3 through HSP26, TPI1, and ATP to GPM1.↪
  The complex ATPGPM1 forms from ATP and GPM1." is converted to ```
HSP26 ->  @ 9.82;
TDH3 + TPI1 -> 2ATP @ k0;
TDH3 -> HSP26 @ k1;
HSP26 -> TPI1 @ k2;
TPI1 -> ATP @ k3;
ATP -> GPM1 @ k4;
ATP + GPM1 -> ATPGPM1 @ k5;
```

Using the conversion rules provided, convert the biochemical reactions↪
  listed below into the formal language. After converting each ↪
 reaction, put them into a code block marked with ```. Inside the code↪
  block, show one reaction per line. No need to provide further ↪
 explanations, just present the model.
````

## E Results on KinModGPT Dataset

See Tables 3–6.

Table 3. Comparison of Mistral$^{7B}_{v0.3}$/LoRA's Output vs. KinModGPT on the "Decay" and "HIV" Examples from [44]

| **Mistral$^{7B}_{v0.3}$/LoRA** | **Expected Target** |
| --- | --- |
| $P \xrightarrow{k_0}$ | $P \xrightarrow{k_0}$ |
| $\xrightarrow{\mathbf{1.0}} \mathbf{P}$ | |
| $M + M \xrightarrow{k_0} E$ | $M + M \xrightarrow{k_0} E$ |
| $E \xrightarrow{k_1} 2M$ | $E \xrightarrow{k_1} 2M$ |
| $E + S \xrightarrow{k_2} ES$ | $E + S \xrightarrow{k_2} ES$ |
| $ES \xrightarrow{k_3} E + S$ | $ES \xrightarrow{k_3} E + S$ |
| $E + P \xrightarrow{k_4} EP$ | $E + P \xrightarrow{k_4} EP$ |
| $EP \xrightarrow{k_5} E + P$ | $EP \xrightarrow{k_5} E + P$ |
| $ES \xrightarrow{k_6} E + P$ | $ES \xrightarrow{k_6} E + P$ |
| $E + I \xrightarrow{k_7} EI$ | $E + I \xrightarrow{k_7} EI$ |
| $EI \xrightarrow{k_8} E + I$ | $EI \xrightarrow{k_8} E + I$ |
| $EI \xrightarrow{k_9} EJ$ | $EI \xrightarrow{k_9} EJ$ |

Note that we adjusted the ground truth to only include rate constants/variables and disregard the kinetic functions. Bolded entries mark differences.

Table 4. Comparison of Mistral$^{7B}_{v0.3}$/LoRA's Output vs. KinModGPT on the "Three-step" Example from [44]

| **Mistral$^{7B}_{v0.3}$/LoRA** | **Expected Target** |
| --- | --- |
| $\mathbf{S \xrightarrow{k_0} M1}$ | $\mathbf{S + E1 \xrightarrow{k_0} M1 + E1}$ |
| $\mathbf{M1 \xrightarrow{k_1} M2}$ | $\mathbf{M1 + E2 \xrightarrow{k_1} M2 + E2}$ |
| $\mathbf{M2 \xrightarrow{k_2} P}$ | $\mathbf{M2 + E3 \xrightarrow{k_2} P + E3}$ |
| $M1 + E1 \xrightarrow{k_3} E1 + M1$ | |
| $M2 + E2 \xrightarrow{k_4} E2 + M2$ | |
| $P + E3 \xrightarrow{k_5} E3 + P$ | |
| $G1 \xrightarrow{k_6} E1$ | $G1 \xrightarrow{k_6} E1$ |
| $G2 \xrightarrow{k_7} E2$ | $G2 \xrightarrow{k_7} E2$ |
| $G3 \xrightarrow{k_8} E3$ | $G3 \xrightarrow{k_8} E3$ |
| $\mathbf{M1 \xrightarrow{k_9} G1}$ | $\mathbf{M1 \xrightarrow{k_3} G1 + M1}$ |
| $\mathbf{M2 \xrightarrow{k_{10}} G2}$ | $\mathbf{M2 \xrightarrow{k_4} G2 + M2}$ |
| $\mathbf{P \xrightarrow{k_{11}} G3}$ | $\mathbf{P \xrightarrow{k_5} G3 + P}$ |
| $E1 \xrightarrow{k_{12}}$ | $E1 \xrightarrow{k_{12}}$ |
| $E2 \xrightarrow{k_{13}}$ | $E2 \xrightarrow{k_{13}}$ |
| $E3 \xrightarrow{k_{14}}$ | $E3 \xrightarrow{k_{14}}$ |
| $G1 \xrightarrow{k_{15}}$ | $G1 \xrightarrow{k_{15}}$ |
| $G2 \xrightarrow{k_{16}}$ | $G2 \xrightarrow{k_{16}}$ |
| $G3 \xrightarrow{k_{17}}$ | $G3 \xrightarrow{k_{17}}$ |
| $\xrightarrow{\mathbf{1.0}} \mathbf{S}$ | |

Note that we adjusted the ground truth to only include rate constants/variables and disregard the kinetic functions. Bolded entries mark differences.

Table 5. Comparison of $\text{Mistral}^{7B}_{v0.3}$/LoRA's Output vs. KinModGPT on the "Heat-shock Response" Example from [44] (First Part)

| **$\text{Mistral}^{7B}_{v0.3}$/LoRA** | **Expected Target** |
|---|---|
| s70 + RNAP $\xrightarrow{k_0}$ s70_RNAP | s70 + RNAP $\xrightarrow{k_0}$ s70_RNAP |
| s70_RNAP $\xrightarrow{k_1}$ s70 + RNAP | s70_RNAP $\xrightarrow{k_1}$ s70 + RNAP |
| Pg + s70_RNAP $\xrightarrow{k_2}$ Pg_s70_RNAP | Pg + s70_RNAP $\xrightarrow{k_2}$ Pg_s70_RNAP |
| Pg_s70_RNAP $\xrightarrow{k_3}$ Pg + s70_RNAP | Pg_s70_RNAP $\xrightarrow{k_3}$ Pg + s70_RNAP |
| RNAP + s32 $\xrightarrow{k_4}$ RNAP_s32 | RNAP + s32 $\xrightarrow{k_4}$ RNAP_s32 |
| RNAP_s32 $\xrightarrow{k_5}$ RNAP + s32 | RNAP_s32 $\xrightarrow{k_5}$ RNAP + s32 |
| Ph + RNAP_s32 $\xrightarrow{k_6}$ Ph_RNAP_s32 | Ph + RNAP_s32 $\xrightarrow{k_6}$ Ph_RNAP_s32 |
| Ph_RNAP_s32 $\xrightarrow{k_7}$ Ph + RNAP_s32 | Ph_RNAP_s32 $\xrightarrow{k_7}$ Ph + RNAP_s32 |
| s32 + DnaK $\xrightarrow{k_8}$ s32_DnaK | s32 + DnaK $\xrightarrow{k_8}$ s32_DnaK |
| | **s32_DnaK $\xrightarrow{k_9}$ s32 + DnaK** |
| s32 + FtsH $\xrightarrow{k_9}$ s32_FtsH | s32 + FtsH $\xrightarrow{k_{10}}$ s32_FtsH |
| | **s32_FtsH $\xrightarrow{k_{11}}$ s32 + FtsH** |
| Punfold + DnaK $\xrightarrow{k_{10}}$ Punfold_DnaK | Punfold + DnaK $\xrightarrow{k_{12}}$ Punfold_DnaK |
| | **Punfold_DnaK $\xrightarrow{k_{13}}$ Punfold + DnaK** |
| D + s70_RNAP $\xrightarrow{k_{11}}$ D_s70_RNAP | D + s70_RNAP $\xrightarrow{k_{14}}$ D_s70_RNAP |
| D_s70_RNAP $\xrightarrow{k_{12}}$ D + s70_RNAP | D_s70_RNAP $\xrightarrow{k_{15}}$ D + s70_RNAP |
| D + RNAP_s32 $\xrightarrow{k_{13}}$ D_RNAP_s32 | D + RNAP_s32 $\xrightarrow{k_{16}}$ D_RNAP_s32 |
| D_RNAP_s32 $\xrightarrow{k_{14}}$ D + RNAP_s32 | D_RNAP_s32 $\xrightarrow{k_{17}}$ D + RNAP_s32 |
| RNAP + D $\xrightarrow{k_{15}}$ RNAP_D | RNAP + D $\xrightarrow{k_{18}}$ RNAP_D |
| RNAP_D $\xrightarrow{k_{16}}$ RNAP + D | RNAP_D $\xrightarrow{k_{19}}$ RNAP + D |
| s32_DnaK + FtsH $\xrightarrow{k_{17}}$ s32_DnaK_FtsH | s32_DnaK + FtsH $\xrightarrow{k_{20}}$ s32_DnaK_FtsH |
| s32_DnaK_FtsH $\xrightarrow{k_{18}}$ s32_DnaK + FtsH | s32_DnaK_FtsH $\xrightarrow{k_{21}}$ s32_DnaK + FtsH |
| s32_FtsH $\xrightarrow{k_{19}}$ FtsH | s32_FtsH $\xrightarrow{k_{22}}$ FtsH |
| s32_DnaK_FtsH $\xrightarrow{k_{20}}$ DnaK + FtsH | s32_DnaK_FtsH $\xrightarrow{k_{23}}$ DnaK + FtsH |
| Pfold $\xrightarrow{k_{21}}$ Punfold | Pfold $\xrightarrow{k_{24}}$ Punfold |
| Punfold_DnaK $\xrightarrow{k_{22}}$ Pfold + DnaK | Punfold_DnaK $\xrightarrow{k_{25}}$ Pfold + DnaK |

Note that we adjusted the ground truth to only include rate constants/variables and disregard the kinetic functions. Bolded entries mark differences.

Table 6. Comparison of Mistral$^{7B}_{v0.3}$/LoRA's Output vs. KinModGPT on the "Heat-shock Response" Example from [44] (Second Part)

| **Mistral$^{7B}_{v0.3}$/LoRA** | **Expected Target** |
| --- | --- |
| **Pg_s70_RNAP $\xrightarrow{k_{23}}$ mRNA_s32** | $\xrightarrow{k_{26}}$ mRNA_s32 |
| **Pg_s70_RNAP $\xrightarrow{k_{24}}$ mRNA_DnaK** | |
| **Pg_s70_RNAP $\xrightarrow{k_{25}}$ mRNA_FtsHi** | |
| **Ph_RNAP_s32 $\xrightarrow{k_{26}}$ mRNA_DnaK** | $\xrightarrow{k_{26}}$ mRNA_DnaK |
| **Ph_RNAP_s32 $\xrightarrow{k_{27}}$ mRNA_FtsH** | $\xrightarrow{k_{28}}$ mRNA_FtsH |
| $\xrightarrow{k_{28}}$ mRNA_Protein | $\xrightarrow{k_{29}}$ mRNA_Protein |
| $\xrightarrow{k_{29}}$ s32 | $\xrightarrow{k_{30}}$ s32 |
| $\xrightarrow{k_{30}}$ FtsH | $\xrightarrow{k_{31}}$ FtsH |
| $\xrightarrow{k_{31}}$ DnaK | $\xrightarrow{k_{32}}$ DnaK |
| $\xrightarrow{k_{32}}$ Pfold | $\xrightarrow{k_{33}}$ Pfold |
| **s32 $\xrightarrow{k_{33}}$ mRNA_s32** | mRNA_s32 $\xrightarrow{k_{34}}$ |
| **DnaK $\xrightarrow{k_{34}}$ mRNA_DnaK** | mRNA_DnaK $\xrightarrow{k_{35}}$ |
| **FtsH $\xrightarrow{k_{35}}$ mRNA_FtsH** | mRNA_FtsH $\xrightarrow{k_{36}}$ |
| **Protein $\xrightarrow{k_{36}}$ mRNA_Protein** | mRNA_Protein $\xrightarrow{k_{37}}$ |
| **s32 $\xrightarrow{k_{38}}$ s32_DnaK** | |
| **s32 $\xrightarrow{k_{39}}$ s32_FtsH** | |
| **s32 $\xrightarrow{k_{40}}$ s32_DnaK_FtsH** | |
| s32 $\xrightarrow{k_{41}}$ | s32 $\xrightarrow{k_{38}}$ |
| s32_DnaK $\xrightarrow{k_{42}}$ | s32_DnaK $\xrightarrow{k_{39}}$ |
| s32_FtsH $\xrightarrow{k_{43}}$ | s32_FtsH $\xrightarrow{k_{40}}$ |
| s32_DnaK_FtsH $\xrightarrow{k_{44}}$ | s32_DnaK_FtsH $\xrightarrow{k_{41}}$ |
| FtsH $\xrightarrow{k_{45}}$ | FtsH $\xrightarrow{k_{42}}$ |
| DnaK $\xrightarrow{k_{46}}$ | DnaK $\xrightarrow{k_{43}}$ |
| Punfold_DnaK $\xrightarrow{k_{47}}$ | Punfold_DnaK $\xrightarrow{k_{44}}$ |
| Pfold $\xrightarrow{k_{48}}$ | Pfold $\xrightarrow{k_{45}}$ |
| Punfold $\xrightarrow{k_{49}}$ | Punfold $\xrightarrow{k_{46}}$ |
| **FtsH $\xrightarrow{k_{50}}$ FtsH** | |
| **DnaK $\xrightarrow{k_{51}}$ DnaK** | |
| **DnaK $\xrightarrow{k_{52}}$ Punfold_DnaK** | |
| **Pfold $\xrightarrow{k_{53}}$ Pfold** | |
| **Pfold $\xrightarrow{k_{54}}$ Punfold** | |
| RNAP_s32 $\xrightarrow{k_{55}}$ RNAP | RNAP_s32 $\xrightarrow{k_{47}}$ RNAP |
| Ph_RNAP_s32 $\xrightarrow{k_{56}}$ Ph + R | Ph_RNAP_s32 $\xrightarrow{k_{48}}$ Ph + RNAP |
| D_RNAP_s32 $\xrightarrow{k_{57}}$ RNAP_D | D_RNAP_s32 $\xrightarrow{k_{49}}$ RNAP_D |

Note that we adjusted the ground truth to only include rate constants/variables and disregard the kinetic functions. Bolded entries mark differences.